\pdfoutput=1

\documentclass[11pt]{article}

\usepackage{amsmath}
\usepackage{mathtools}
\usepackage{amssymb}
\usepackage{times}
\usepackage{latexsym}
\usepackage{booktabs}
\usepackage{multirow}
\usepackage{adjustbox}
\usepackage{siunitx}
\usepackage{graphicx}
\usepackage{authblk}
\usepackage{float}

\usepackage{array}
\newcolumntype{P}[1]{>{\centering\arraybackslash}p{#1}}

\usepackage[]{ACL2023}

\usepackage{times}
\usepackage{latexsym}

\usepackage[T1]{fontenc}

\usepackage[utf8]{inputenc}

\usepackage{microtype}

\usepackage{inconsolata}

\title{Using Bottleneck Adapters to Identify Cancer in Clinical Notes under Low-Resource Constraints}

\author{
  Omid Rohanian$^{1,6}$,
  Hannah Jauncey$^{3}$,
  Mohammadmahdi Nouriborji$^{5,6}$,
  Vinod Kumar Chauhan$^{1}$,\\
  Bronner P. Gonçalves$^{2}$,
  Christiana Kartsonaki$^{2}$,
  ISARIC Clinical Characterisation Group$^{2}\dagger$,
  Laura Merson$^{2}$,\\
  David Clifton$^{1,4}$\\
  $^1$Department of Engineering Science, University of Oxford, Oxford, UK \\
  $^2$ISARIC, Pandemic Sciences Institute, University of Oxford, Oxford, UK \\
  $^3$ Infectious Diseases Data Observatory (IDDO), University of Oxford, UK\\
  $^4$Oxford-Suzhou Centre for Advanced Research, Suzhou, China \\
  $^5$Sharif University of Technology, Tehran, Iran\\
  $^6$NLPie Research, Oxford, UK \\\\

  \texttt{\{omid.rohanian,david.clifton,vinod.kumar\}@eng.ox.ac.uk}\\
  \texttt{\{hannah.jauncey,laura.merson,bronner.goncalves\}@ndm.ox.ac.uk}\\
  \texttt{m.nouriborji@nlpie.com}\\
  \texttt{christiana.kartsonaki@dph.ox.ac.uk}
 }

\begin{document}
\maketitle
\def\thefootnote{$\dagger$}\footnotetext{Please refer to Appendix \ref{isaric-names} for the full list of collaborators.}  
\def\thefootnote{\arabic{footnote}}
\begin{abstract}
Processing information locked within clinical health records is a challenging task that remains an active area of research in biomedical NLP.  In this work, we evaluate a broad set of machine learning techniques ranging from simple RNNs to specialised transformers such as BioBERT on a dataset containing clinical notes along with a set of annotations indicating whether a sample is cancer-related or not. 

Furthermore, we specifically employ efficient fine-tuning methods from NLP, namely, bottleneck adapters and prompt tuning, to adapt the models to our specialised task. Our evaluations suggest that fine-tuning a frozen BERT model pre-trained on natural language and with bottleneck adapters outperforms all other strategies, including full fine-tuning of the specialised BioBERT model. Based on our findings, we suggest that using bottleneck adapters in low-resource situations with limited access to labelled data or processing capacity could be a viable strategy in biomedical text mining. The code used in the experiments are going to be made available at [LINK ANONYMIZED].
\end{abstract}

\section{Introduction}
\label{intro}

Clinical notes involve important information about patients and their current state and medical history. Automatic processing of these notes and the terms that appear in them would help researchers classify them into standard conditions that can also be looked up in medical knowledge-bases.   
In combination with other medical signals, this information has been shown to be useful in predicting in-hospital mortality rate \citep{deznabi-etal-2021-predicting}, prolonged mechanical ventilation \citep{huang-etal-2020-clinical}, or clinical outcome \citep{van-aken-etal-2021-clinical}, among others. 

In this work, we looked at a real clinical notes database and designed a pilot experiment in which a set of different ML models were used to predict whether a clinical note is cancer-related or not. The incentive behind this experiment is to help clinicians and data curators to automatically search for and identify notes that signal a particular medical condition, instead of solely relying on laborious human annotation and keyword-based search. 

The promise of ML is in automating this task reasonably close to human-level performance and ultimately expanding this work to include other conditions in a multi-class scenario. Ideally a model would be able to identify cancer types that are not seen during training and would be able to have some understanding of context and grammar to be sensitive to negation.

\subsection*{Contributions} In this work, we targeted the task of disease identification within a clinical notes dataset. We tested a range of different models including RNN-based and transformer-based architectures to tackle this problem. We particularly focused on efficient fine-tuning approaches to adapt our pre-trained models to the biomedical task. The novelty of this work is in the successful application of bottleneck adapters to the cancer identification task which to the best of our knowledge has not been explored before.We compare this method with multiple other strong baselines and conduct experiments and analyses to evaluate these different approaches. The systems developed in this study and those that will follow in related future work will be added to the data curation system of a biomedical database with the aim to enable automatic processing of clinical notes in real EHR data. 

\section{Pre-Trained Transformers and Fine-Tuning}

In recent years, the Transformers architecture \citep{vaswani2017attention} and large language models (LMs) have become the staple baseline for many NLP tasks. The conventional paradigm is to first pre-train an LM on a large corpus of general text (e.g. Wikipedia) with a pre-training objective such as masked or causal language modeling and then fine-tune the LM on downstream tasks.

In our task, we focus on transformers pre-trained with the Masked Language Modeling (MLM) objective. In MLM, a portion of the text is masked out and the objective of the model is to learn to reconstruct the masked portion based on the available context. The most commonly used model pre-trained with MLM is named BERT \citep{devlin-etal-2019-bert}.

Despite BERT's promising results on many downstream NLP tasks, it has been shown that large LMs pre-trained on generic text do not always perform well on specialised domains like biomedical tasks \citep{lee2020biobert,gururangan2020don}. The standard approach, therefore, is to pre-train models on corpora that are related to the target domain. BioBERT \citep{lee2020biobert} is an example of an LM trained on specialized data. It is trained on a large corpus of general and biomedical texts making it a strong model for biomedical text mining.

\subsection{Efficient Fine-Tuning Methods}

The benefits of fine-tuning large LMs for downstream applications are offset by a significant computational cost. Some LMs, for example, include more than $100$ billion parameters, making their fine-tuning costly. Furthermore, complete fine-tuning may be ineffective when the amount of training data is small or different from the initial domain that the model was trained on, which might result in catastrophic forgetting.

As a response to these limitations, more efficient fine-tuning approaches have been developed, among which prompt tuning (\ref{pt}) and bottleneck adapters (\ref{ba}) are two of the most effective and well-known. 

\subsection{Bottleneck Adapters}
\label{ba}

Bottleneck Adapters (BAs) \citep{adapter,pfeiffer-etal-2021-adapterfusion,adapter_drop,adapter_hub} are Multi-layer Perceptron (MLP) blocks that are made up of a down-projection dense layer, an activation function, and an up-projection dense layer with a residual connection. These blocks are inserted between the frozen attention and feed-forward blocks of a pre-trained LM, and only these modules will be updated during fine-tuning. This method has proven to be effective in terms of both computational and parameter efficiency.  

\citet{adapter} showed that by training only around $3$\% of the parameters, BERT trained with adapters can get competitive results compared to complete fine-tuning. Adapter tuning can be expressed in the below equation where $X_i$ is the output of the frozen attention or MLP component of the $i_{th}$ layer of the pre-trained LM.
\begin{equation}
    O_i = f_{up}(Activation(f_{down}(X_i))) + X_i
\end{equation}

\subsection{Prompt Tuning}
\label{pt}

Another efficient method of fine-tuning is called Prompt Tuning (PT) \citep{prefix_tuning,prompt_tuning}. PT is mostly used for autoregressive LMs such as GPT \citep{gpt-3}. In this approach, a set of learnable vectors (prompt) are concatenated with the original input and passed to the LM. During fine-tuning, the objective is to learn a prompt which is intended to encode task-specific knowledge for the downstream task while the original model parameters are kept frozen. In some variations of PT, instead of concatenating a set of learnable vectors with the input before passing it to the model once, a set of prompts are learned for each individual attention layer of the pre-trained LM \citep{prefix_tuning}. The PT approach used in this study can be expressed in the below equation where $Attention_i$ is the attention block of the $i_{th}$ layer of the pre-trained transformer and $P^k_i$ and $P^v_i$ denote the learnable prompts for keys and values respectively.
\begin{equation}
    O_i = Attention_{i}(Q_{i}, [P^{k}_{i} , K_i] , [P^{v}_{i} , V_i])
\end{equation}

\subsection{Bottleneck Adapters in Biomedical Domain}
BAs are increasingly used for efficient knowledge extraction and domain adaptation due to their parameter efficiency and low computational cost. Following this trend, there are some works in the biomedical domain that have used adapters to insert task-specific knowledge via pre-training into the LMs \citep{grover2021multi,lu2021parameter}, or employed them in layer adaptation for developing compact biomedical models \citep{nouriborji2022minialbert}.

\section{Challenges of Identifying Cancer-Related Records}
Clinical notes usually involve abbreviated and non-standard language. A single concept like cancer is mentioned in different ways depending on cancer subtype. The same subtype might have a scientific and a commonly known variant and both can appear in the text. Grammar is sometimes broken and language can appear cryptic. Another issue is the prevalence of misspellings which further complicates this task. 

There are also words that co-occur with a condition and can easily confound the model. For instance, words like `breast' and `lung' which are not specific to cancer appear a lot in cancer-related samples and the model can mistake them for a cancer signal. Another important issue is negation. If a condition is ruled out, ideally a model should not return positive. However since most rows that are classed as positive in the dataset include the token `cancer', an example like `not cancer' could be mistaken as positive. Encoding awareness of negation into the model is a challenge since it is known that pre-trained LMs lack an innate ability to handle negation \citep{hosseini2021understanding}.

\section{Dataset and Annotation}
The dataset in this pilot experiment was provided by ISARIC, a global initiative that, among other things, provides tools and resources to facilitate clinical research\footnote{The ISARIC COVID-19 Data Platform is a global partnership of more than $1,700$ institutions across more than $60$ countries. Accreditation of the individuals, institutions and funders that contributed to this effort can be found in the supplementary material. These partners have combined data and expertise to accelerate the pandemic response and improve patient outcomes. For more information on ISARIC, see \url{https://isaric.org}. Data are available for access via application to the Data Access Committee at \url{www.iddo.org/covid-19}. }.  The larger dataset contains $125381$ rows corresponding to clinical notes related to different conditions and patients. For the purposes of this experiment a portion of this data was annotated for presence of cancer. The annotated subset contains $2563$ rows that include cancer labels, out of which $343$ are repeated notes where the doctors have written the same cancer-related note for a different patient. The human experts who tagged the data for cancer, had access to a set of cancer-related terms to guide them in the annotation. The negative cohort of 3K rows was generated by filtering out the larger data by any row that contained keywords that could potentially signal cancer definitively or with a very high possibility. The details of the lists and more information on the annotation scheme are included in the appendix (\ref{annot}).   
\section{Experiments}
The experiments in this work are divided into two categories, namely, attention-based and RNN-based methods. We conducted all our experiments on an internal cancer detection dataset with \textasciitilde$6k$ labeled samples with roughly equal instances in each classes and evaluated them on a gold standard consisting of 1k samples, $31$ of which were positive and the rest negative. Note the distributional shift between training and test sets which reflect the real clinical setting under which the models are expected to perform.

\begin{figure}[t!]
\centering
\includegraphics[width=0.4\textwidth]{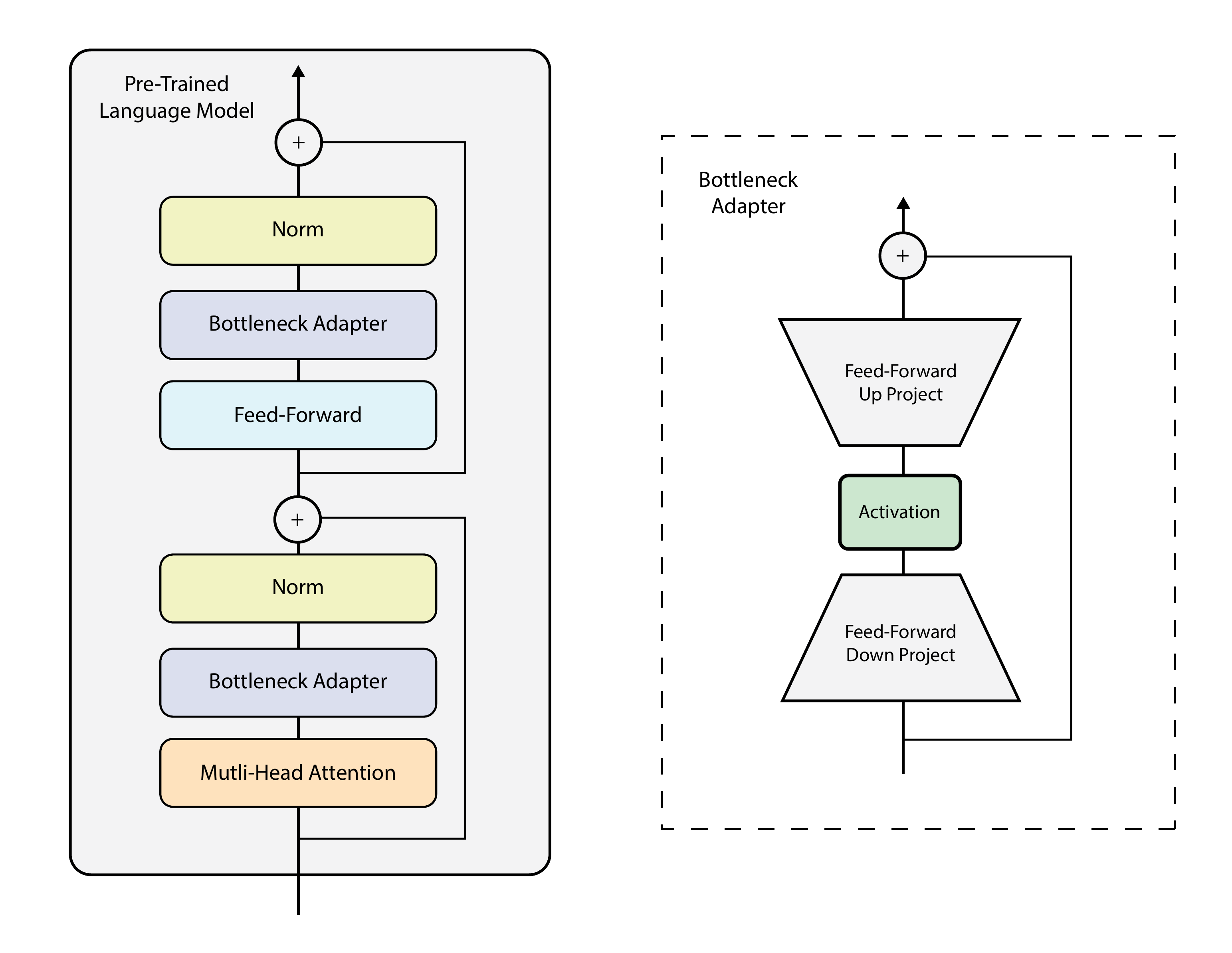}
\caption{The overall architecture of Adapter Tuning, Note that the original parameters of the pre-trained model are kept frozen for the fine-tuning of the model and only the Bottleneck Adapters in between attention and feed-forward layers will be updated.}
\label{model-architecture}
\end{figure}

\subsection{Baselines}
We used three baselines in this work all of which are RNN-based. The initial weights in embedding layer of all the baselines comes from \citet{chen2019biosentvec} which is a \texttt{word2vec} model pre-trained on medical data. The first model is a simple Bi-LSTM, the second uses a 1D-convolution before the Bi-LSTM (CNN-Bi-LSTM), and the final model adds a multi-head self-attention layer after the CNN-Bi-LSTM model (CNN-Bi-LSTM-Att). All models are trained for $24$ epochs with a batch size of $64$.

\begin{table*}[!ht]
    \centering
    \resizebox{10cm}{!}{
    \begin{tabular}{P{3.5cm}|p{3.5cm}|P{1.5cm}|P{1.5cm}|P{2cm}}
        \toprule[1pt]
        Model Architecture & Approach & Precision & Recall & F-Score \\\midrule[1pt]
                 & Bi-LSTM & 0.98 & 0.75 & 0.83\\
        RNN      & CNN-Bi-LSTM & 0.96 & 0.70 & 0.77\\
                 & CNN-Bi-LSTM-Att & 0.96 & 0.72 & 0.79\\
                    \cmidrule[0.8pt]{1-5}
                  & Complete Fine-Tuning & 0.97 & 0.77 & 0.84\\
        BERT      & Adapter-Tuning & \textbf{1.00} & 0.94 & \textbf{0.97}\\
                  & Prompt-Tuning & 0.97 & 0.79 & 0.86\\
                    \cmidrule[0.8pt]{1-5}
                  & Complete Fine-Tuning & 0.99 & 0.84 & 0.90\\
        BioBERT   & Adapter-Tuning & 0.98 & 0.85 & 0.90\\
                  & Prompt-Tuning & 0.98 & 0.87 & 0.92\\
                    \cmidrule[0.8pt]{1-5}
        BERT + Negation      & Adapter-Tuning-500 & 0.98 & \textbf{0.95} & \textbf{0.97}\\
                             & Adapter-Tuning-250 & 0.98 & 0.93 & 0.95\\
                    \bottomrule
    \end{tabular}}
    \caption{\label{t:results} Results obtained on the gold standard dataset with 1k annotated samples. Note that the Adapter-Tuning-500 and Adapter-Tuning-250 denote models trained with 500 and 250 artificially generated negative samples, respectively.}
\end{table*}
\subsection{Approach}
Our aim was to improve upon the strong RNN baselines by the use of efficient fine-tuning of pre-trained transformers, namely, BERT \citep{devlin-etal-2019-bert} and BioBERT \citep{lee2020biobert}. Three fine-tuning approaches were tried: full fine-tuning, tuning with BAs (Sec. \ref{ba}), and PT (Sec. \ref{pt}). 

\subsubsection{Tuning with Adapters}
The BA used in this work is from \citet{adapter} and implemented using Adapter Hub \citep{adapter_hub}. The reduction factor of the adapter is set to $16$ and its activation function is ReLU. The adapters are used after attention layers and feed-forward layers of each transformer block while the parameters of the model are kept frozen. The overall architecture of the model used in this work is depicted in Figure \ref{model-architecture}. 

\subsubsection{Tuning with Prompts}
For the PT, the approach from \citep{prefix_tuning} with a prompt size of $30$ is used and implemented with the Adapter Hub library \citep{adapter_hub}. In this approach, a set of prompts are learned for each attention layer of the frozen language model.

\subsubsection{Encoding knowledge of Negation and Uncertainty}
\label{negation}
Negation is not by default understood by any of the models we have explored in this work. For instance, the phrases `Evidence of lung cancer' and `No Evidence of lung cancer' are both predicted as cancer-related by a model as the negative samples in the training set do not include negation patterns for cancer. To encode some understanding of negation into the model, we analysed the larger dataset and identified a number of different ways a condition can be ruled out with varying degrees of certainty (full list available at \ref{negation-patterns}). We used these examples to generate synthetic negative samples that include a cancer-related term (e.g. `not lymphoma' or `Melanoma not formally diagnosed').   

\section{Results}

Reported results in Table \ref{t:results} are best out of three subsequent runs. For each approach, the hyperparameters that seemed to work best during training were kept fixed for all the runs. Full fine-tuning was done with $5$ epochs and a learning rate of $2e-5$. Tuning with BAs was done with $10$ epochs and a learning rate of $1e-3$. PT was used with $10$ epochs and a learning rate of $1e-4$. All approaches used a batch size of $64$, AdamW Optimizer, Weight Decay of $0.01$, and a cosine scheduler. As can be seen, the best performing model is the BERT trained with Adapters (including variants which are equipped with some notion of negation as explained in \ref{negation}).

Analysing the outputs of individual models, we found that the majority of positive labels in the test set are correctly identified by most models. The bottleneck, however, is the false positives that happen due to the presence of certain words (e.g. `diagnosed with', `lung', `breast' etc) that co-occur with cancer and can cause models to incorrectly label an instance as positive. The best model had only $4$ false positives and no false negatives. The values for the confusion matrices of all the models are provided in \ref{confusion-mat}. 

To alleviate the false positive issue, using the method explained in \ref{negation}, we trained our best model (BERT with adapter-tuning) with additional $250$ and $500$ generated negative samples. The model was subsequently able to predict cases such as `neither cancer nor covid', `lung infection but no cancer', and `diagnosed with covid but not cancer' correctly with only minor performance drops.

A point of strength in all the models was their ability to correctly identify cancer, given rare cancer types that had not occurred in the training set. This generalisation to unseen cancer types indicates that the models can effectively use information from the pre-trained resources they rely upon.     

\section{Conclusion}
In this work, we trained and tested a number of classification approaches as part of a preliminary experiment on a dataset of clinical notes annotated for presence of cancer. We compared a number of RNN models utilising pre-trained biomedical embeddings with two different pre-trained transformer-based models that were fine-tuned in separate ways. We also addressed the issue of negation by integrating negation patterns into the negative training samples. Our findings suggest that efficient fine-tuning of transformers using adapters resulted in the best performance among all the tested approaches. The codes for the experiments conducted in this project are publicly available at \url{https://github.com/omidrohanian/bottleneck-adapters}.

\section*{Limitations}

This work has certain limitations in terms of the scope of the experiments and what can be reliably inferred from them. Our dataset contains notes that are predominantly from anglophone countries. However, there are less than $20\%$ of the rows that originate from non-English speaking regions. They might contain words in other languages (e.g. Italian), and although the disease names are usually rendered similarly as English, our models are pre-trained on English and their ability to process other languages is therefore limited. 

Another issue is the relatively short length of these notes. While some notes span a few sentences, most are very short and no more than $4-5$ tokens in length. This hampers the ability of a contextualised model to derive meaning from the context around each word and limits the power of attention-based architectures that are well-suited for larger contexts. 

In this preliminary study we only targeted one condition and looked at binary classification. The natural step towards a more inclusive experiment would be to consider other conditions and also use multi-class classification setups where a more fine-grained scheme is used to classify a condition. Creating a sizable multi-class and multi-label corpus is a labour-intensive endeavor that requires more time and effort and would be a goal for a future work. We did have access to multi-class annotations for our current training set, however, one major issue is that the cancer-positive cases are a small percentage of the entire rows and among the cancer types themselves, there are types that occur only once or twice and the rest belong to more frequent classes. This would make it harder for the model to learn infrequent classes. We plan to augment the annotations over time to be able to conduct experiments in scenarios beyond binary classification and cancer alone. 

The issue of negation was further complicated in this work by a few cases where the note had been classified as cancer positive because the doctor had identified a history of this condition in the patient but had ruled out or downplayed the possibility of cancer at the present time. Distinguishing a current co-morbidity of cancer from a past history of cancer would introduce further complexity and this work does not attempt to address that.  

\section*{Ethics Statement}
Ethics Committee approval for the data collection and analysis for this work was given by the World Health Organisation Ethics Review Committee (RPC571 and RPC572 on 25 April 2013). National and/or institutional ethics committee approval was additionally obtained by participating sites according to local requirements. 

This work is a part of a global effort to accelerate and improve the collection and analysis of data in the context of infectious disease outbreaks. Rapid characterisation of novel infections is critical to an effective public health response. The model developed will be implemented in data aggregation and curation platforms for outbreak response – supporting the understanding of the variety of data collected by frontline responders. The challenges of implementing robust data collection efforts in a health emergency often result in non-standard data using a wide range of terms. This is especially the case in lower-resourced settings where data infrastructure is lacking. This work aims to improve data processing, and will especially contribute to lower-resource settings to improve health equity. 

\section*{Funding and Acknowledgements}

This work was made possible with the support of UK Foreign, Commonwealth and Development Office and Wellcome [225288/Z/22/Z].

This work was made possible with the support of UK Foreign, Commonwealth and Development Office and Wellcome [225288/Z/22/Z]. Collection of data for the ISARIC Clinical Notes was made possible with the support of UK Foreign, Commonwealth and Development Office and Wellcome [215091/Z/18/Z, 222410/Z/21/Z, 225288/Z/22/Z, 220757/Z/20/Z and 222048/Z/20/Z] and the Bill \& Melinda Gates Foundation [OPP1209135]; CIHR Coronavirus Rapid Research Funding Opportunity OV2170359 and was coordinated out of Sunnybrook Research Institute; was supported by endorsement of the Irish Critical Care- Clinical Trials Group, co-ordinated in Ireland by the Irish Critical Care- Clinical Trials Network at University College Dublin and funded by the Health Research Board of Ireland [CTN-2014-12]; grants from Rapid European COVID-19 Emergency Response research (RECOVER) [H2020 project 101003589] and European Clinical Research Alliance on Infectious Diseases (ECRAID) [965313]; Cambridge NIHR Biomedical Research Centre; Wellcome Trust fellowship [205228/Z/16/Z] and the National Institute for Health Research Health Protection Research Unit (HPRU) in Emerging and Zoonotic Infections (NIHR200907) at the University of Liverpool in partnership with Public Health England (PHE), in collaboration with Liverpool School of Tropical Medicine and the University of Oxford; The dedication and hard work of the Norwegian SARS-CoV-2 study team. 

Research Council of Norway grant no 312780, and a philanthropic donation from Vivaldi Invest A/S owned by Jon Stephenson von Tetzchner; PJMO is supported by the UK’s National Institute for Health Research (NIHR) via Imperial’s Biomedical Research Centre (NIHR Imperial BRC), Imperial’s Health Protection Research Unit in Respiratory Infections (NIHR HPRU RI), the Comprehensive Local Research Networks (CLRNs) and is an NIHR Senior Investigator (NIHR201385); Innovative Medicines Initiative Joint Undertaking under Grant Agreement No. 115523 COMBACTE, resources of which are composed of financial contribution from the European Union’s Seventh Framework Programme (FP7/2007- 2013) and EFPIA companies, in-kind contribution; Stiftungsfonds zur Förderung der Bekämpfung der Tuberkulose und anderer Lungenkrankheiten of the City of Vienna; Project Number: APCOV22BGM; Australian Department of Health grant (3273191); Gender Equity Strategic Fund at University of Queensland, Artificial Intelligence for Pandemics (A14PAN) at University of Queensland, The Australian Research Council Centre of Excellence for Engineered Quantum Systems (EQUS, CE170100009), The Prince Charles Hospital Foundation, Australia; grants from Instituto de Salud Carlos III, Ministerio de Ciencia, Spain; Brazil, National Council for Scientific and Technological Development Scholarship number 303953/2018-7; the Firland Foundation, Shoreline, Washington, USA;  The French COVID cohort (NCT04262921) is sponsored by INSERM and is funding by the REACTing (REsearch \& ACtion emergING infectious diseases) consortium and by a grant of the French Ministry of Health (PHRC n°20-0424);  the South Eastern Norway Health Authority and the Research Council of Norway; and a grant from the Oxford University COVID-19 Research Response fund (grant 0009109); Institute for Clinical Research (ICR), National Institutes of Health (NIH) supported by the Ministry of Health Malaysia; a grant from foundation Bevordering Onderzoek Franciscus.

The investigators acknowledge the philanthropic support of the donors to the University of Oxford’s COVID-19 Research Response Fund; COVID clinical management team, AIIMS, Rishikesh, India; COVID-19 Clinical Management team, Manipal Hospital Whitefield, Bengaluru, India; Italian Ministry of Health ``Fondi Ricerca corrente–L1P6'' to IRCCS Ospedale Sacro Cuore–Don Calabria; and Preparedness work conducted by the Short Period Incidence Study of Severe Acute Respiratory Infection; The dedication and hard work of the Groote Schuur Hospital Covid ICU Team, supported by the Groote Schuur nursing and University of Cape Town registrar bodies coordinated by the Division of Critical Care at the University of Cape Town.

This work uses data provided by patients and collected by the NHS as part of their care and support \#DataSavesLives. The data used for this research were obtained from ISARIC4C. We are extremely grateful to the 2648 frontline NHS clinical and research staff and volunteer medical students who collected these data in challenging circumstances; and the generosity of the patients and their families for their individual contributions in these difficult times. The COVID-19 Clinical Information Network (CO-CIN) data was collated by ISARIC4C Investigators. Data and Material provision was supported by grants from: the National Institute for Health Research (NIHR; award CO-CIN-01), the Medical Research Council (MRC; grant MC\_PC\_19059), and by the NIHR Health Protection Research Unit (HPRU) in Emerging and Zoonotic Infections at University of Liverpool in partnership with Public Health England (PHE), (award 200907), NIHR HPRU in Respiratory Infections at Imperial College London with PHE (award 200927), Liverpool Experimental Cancer Medicine Centre (grant C18616/A25153), NIHR Biomedical Research Centre at Imperial College London (award ISBRC-1215-20013), and NIHR Clinical Research Network providing infrastructure support. We also acknowledge the support of Jeremy J Farrar and Nahoko Shindo.

This work was supported in part by the National Institute for Health Research (NIHR) Oxford Biomedical Research Centre (BRC), and in part by an InnoHK Project at the Hong Kong Centre for Cerebro-cardiovascular Health Engineering (COCHE). OR acknowledges the support of the Medical Research Council (grant number MR/W01761X/). DAC was supported by an NIHR Research Professorship, an RAEng Research Chair, COCHE, and the Pandemic Sciences Institute at the University of Oxford. The views expressed are those of the authors and not necessarily those of the NHS, NIHR, MRC, COCHE, or the University of Oxford.

\bibliography{anthology,custom}
\bibliographystyle{acl_natbib}

\appendix

\section{Appendix}
\label{sec:appendix}

\subsection{ISARIC Dataset and the Annotation Procedure}
\label{annot}
As of January 2022, the ISARIC COVID-19 Clinical Database comprises of standardised data from over 800,601 hospitalised COVID-19 patients, collated during the pandemic to facilitate high quality and timely research \citep{group2021value}. The database contains demographic and clinical data, including hospital admission and discharge records, signs and symptoms, comorbidities, vital signs, treatments, and outcomes. Cancer is one of many comorbidities that has been found to be relevant to patient outcomes and was therefore chosen as the focus of this work \citep{palmieri2020cancer}. For initial model development in this experiment, a stratified sample of non-prespecified (free text) medical terms from the ISARIC COVID-19 Clinical Database was extracted. The sample was searched by a data manger (HJ, with previous clinical medicine experience) for the following cancer-related terms \footnote{* Denotes terms for which there was uncertainty as to the nature of the neoplasm/diagnosis; these terms were labelled as cancer-related during this process.}: 

\begin{itemize}
    \item Adenocarcinoma 
    \item Adeno-carcinoma 
    \item Adeno CA  
    \item Adenocarcinome 
    \item ALL  
    \item AML 
    \item Astrocytoma* 
    \item BCC 
    \item Blastoma 
    \item CA
    \item Cancer 
    \item Carcinoma 
    \item Carcinosarcoma 
    \item Carcinoid* 
    \item Cholangiocarcinoma 
    \item CLL 
    \item CML 
    \item DLBCL 
    \item Ependymoma* 
    \item Ewing 
    \item Gastrointestinal stromal tumour* 
    \item Gestational trophoblastic disease* 
    \item GBM (Glioblastoma multiforme) 
    \item Glioblastoma 
    \item GIST*
    \item HCC  
    \item Hodgkin 
    \item Hepatoblastoma
    \item Kaposi 
    \item Leukaemia  
    \item Leukemia 
    \item Lymphoma 
    \item Malignancy 
    \item Malignant  
    \item MDS  
    \item Melanoma  
    \item Meningioma* 
    \item Mesothelioma  
    \item Met  
    \item Metastases  
    \item Metastatic  
    \item Mets  
    \item Mycosis fungoides  
    \item Myelodysplasia 
    \item MDS 
    \item Myelodysplastic syndrome 
    \item Myeloma  
    \item NHL  
    \item NSCLC  
    \item Neuroblastoma  
    \item Neuroendocrine cancer
    \item Neuroendocrine neoplasm* 
    \item Neuroendocrine tumour* 
    \item Non-Hodgkin  
    \item Osteosarcoma  
    \item Pancoast  
    \item Retinoblastoma  
    \item Sarcoma  
    \item SCC  
    \item SCLC  
    \item Sezary  
    \item TCC  
    \item Tumor* 
    \item Tumour* 
    \item Wilms
\end{itemize}

\subsection{Negation Patterns}
\label{negation-patterns}

Having observed general patterns of negation in the larger dataset, we used the following $12$ templates to generate negated cases of cancer:

\begin{enumerate}
    \item Most likely not \texttt{CONDITION}. 
    \item Not \texttt{CONDITION}.
    \item \texttt{CONDITION} is ruled out.
    \item No \texttt{CONDITION} was observed 
    \item Unlikely to be \texttt{CONDITION}. 
    \item Not suggestive of \texttt{CONDITION}. 
    \item No indication of \texttt{CONDITION}. 
    \item No \texttt{CONDITION} detected 
    \item \texttt{CONDITION} not diagnosed.
    \item \texttt{CONDITION} not confirmed. 
    \item No evidence of \texttt{CONDITION}. 
    \item No \texttt{CONDITION} found. 
\end{enumerate}

Below are some generated examples:

\begin{itemize}
    \item Lung cancer is ruled out.
    \item No Gastrointestinal Stromal tumour was observed. 
    \item Unlikely to be malignant.
    \item Not suggestive of CA. 
    \item No malignancy detected. 
    \item CA not diagnosed. 
    \item Cancer not formally diagnosed by a doctor. 
\end{itemize}

\subsection{Confusion Matrices of Tested Models}
\label{confusion-mat}

Table \ref{t:results2} contains the confusion matrices for all the classification models that are compared in this work. 

\begin{table*}[!ht]
    \centering{
    \begin{tabular}{P{3.5cm}|p{3.5cm}|P{1.5cm}P{1.5cm}P{1.5cm}P{1.5cm}}
        \toprule[1pt]
        Model Architecture & Approach & TN & FP & FN & TP \\\midrule[1pt]
                 & Bi-LSTM & 939 & 30 & 0 & 31\\
        RNN      & CNN-Bi-LSTM & 925 & 44 & 1 & 30\\
                 & CNN-Bi-LSTM-Att & 931 & 38 & 1 & 30\\
                    \cmidrule[0.8pt]{1-6}
                  & Complete Fine-Tuning & 944 & 25 & 1 & 30\\
        BERT      & Adapter-Tuning & 965 & 4 & 0 & 31\\
                  & Prompt-Tuning & 947 & 22 & 1 & 30\\
                    \cmidrule[0.8pt]{1-6}
                  & Complete Fine-Tuning & 954 & 15 & 0 & 31\\
        BioBERT   & Adapter-Tuning & 956 & 13 & 1 & 30\\
                  & Prompt-Tuning & 959 & 10 & 1 & 30\\
                    \cmidrule[0.8pt]{1-6}
        BERT + Negation      & Adapter-Tuning-500  & 966 & 3 & 1 & 30\\
                             & Adapter-Tuning-250  & 964 & 5 & 1 & 30\\
                    \bottomrule
    \end{tabular}}
    \caption{\label{t:results2} Confusion matrices obtained from the gold standard dataset with 1k annotated samples. Adapter-Tuning-500 and Adapter-Tuning-250 denote models trained with 500 and 250 artificially generated negative samples, respectively.}
\end{table*}

\subsection{ISARIC Clinical Characterisation Group}
\label{isaric-names}

Ali Abbas, Sheryl Ann Abdukahil, Nurul Najmee Abdulkadir, Ryuzo Abe, Laurent Abel, Amal Abrous, Lara Absil, Kamal Abu Jabal, Nashat Abu Salah, Subhash Acharya, Andrew Acker, Shingo Adachi, Elisabeth Adam, Francisca Adewhajah, Enrico Adriano, Diana Adrião, Saleh Al Ageel, Shakeel Ahmed, Marina Aiello, Kate Ainscough, Eka Airlangga, Tharwat Aisa, Ali Ait Hssain, Younes Ait Tamlihat, Takako Akimoto, Ernita Akmal, Eman Al Qasim, Razi Alalqam, Aliya Mohammed Alameen, Angela Alberti, Tala Al-dabbous, Senthilkumar Alegesan, Cynthia Alegre, Marta Alessi, Beatrice Alex, Kévin Alexandre, Abdulrahman Al-Fares, Huda Alfoudri, Adam Ali, Imran Ali, Naseem Ali Shah, Kazali Enagnon Alidjnou, Jeffrey Aliudin, Qabas Alkhafajee, Clotilde Allavena, Nathalie Allou, Aneela Altaf, João Alves, João Melo Alves, Rita Alves, Joana Alves Cabrita, Maria Amaral, Nur Amira, Heidi Ammerlaan, Phoebe Ampaw, Roberto Andini, Claire Andréjak, Andrea Angheben, François Angoulvant, Sophia Ankrah, Séverine Ansart, Sivanesen Anthonidass, Massimo Antonelli, Carlos Alexandre Antunes de Brito, Kazi Rubayet Anwar, Ardiyan Apriyana, Yaseen Arabi, Irene Aragao, Francisco Arancibia, Carolline Araujo, Antonio Arcadipane, Patrick Archambault, Lukas Arenz, Jean-Benoît Arlet, Christel Arnold-Day, Ana Aroca, Lovkesh Arora, Rakesh Arora, Elise Artaud-Macari, Diptesh Aryal, Motohiro Asaki, Angel Asensio, Elizabeth A. Ashley, Muhammad Ashraf, Namra Asif, Mohammad Asim, Jean Baptiste Assie, Amirul Asyraf, Minahel Atif, Anika Atique, AM Udara Lakshan Attanyake, Johann Auchabie, Hugues Aumaitre, Adrien Auvet, Eyvind W. Axelsen, Laurène Azemar, Cecile Azoulay, Benjamin Bach, Delphine Bachelet, Claudine Badr, Roar Bævre-Jensen, Nadia Baig, J. Kenneth Baillie, J Kevin Baird, Erica Bak, Agamemnon Bakakos, Nazreen Abu Bakar, Andriy Bal, Mohanaprasanth Balakrishnan, Valeria Balan, Irene Bandoh, Firouzé Bani-Sadr, Renata Barbalho, Nicholas Yuri Barbosa, Wendy S. Barclay, Saef Umar Barnett, Michaela Barnikel, Helena Barrasa, Audrey Barrelet, Cleide Barrigoto, Marie Bartoli, Cheryl Bartone, Joaquín Baruch, Mustehan Bashir, Romain Basmaci, Muhammad Fadhli Hassin Basri, Denise Battaglini, Jules Bauer, Diego Fernando Bautista Rincon, Denisse Bazan Dow, Abigail Beane, Alexandra Bedossa, Ker Hong Bee, Netta Beer, Husna Begum, Sylvie Behilill, Karine Beiruti, Albertus Beishuizen, Aleksandr Beljantsev, David Bellemare, Anna Beltrame, Beatriz Amorim Beltrão, Marine Beluze, Nicolas Benech, Lionel Eric Benjiman, Dehbia Benkerrou, Suzanne Bennett, Binny Benny, Luís Bento, Jan-Erik Berdal, Delphine Bergeaud, Hazel Bergin, José Luis Bernal Sobrino, Giulia Bertoli, Lorenzo Bertolino, Simon Bessis, Adam Betz, Sybille Bevilcaqua, Karine Bezulier, Amar Bhatt, Krishna Bhavsar, Isabella Bianchi, Claudia Bianco, Farah Nadiah Bidin, Moirangthem Bikram Singh, Felwa Bin Humaid, Mohd Nazlin Bin Kamarudin, François Bissuel, Patrick Biston, Laurent Bitker, Jonathan Bitton, Pablo Blanco-Schweizer, Catherine Blier, Frank Bloos, Mathieu Blot, Lucille Blumberg, Filomena Boccia, Laetitia Bodenes, Debby Bogaert, Anne-Hélène Boivin, Isabela  Bolaños, Pierre-Adrien Bolze, François Bompart, Patrizia Bonelli, Aurelius Bonfasius, Joe Bonney, Diogo Borges, Raphaël Borie, Hans Martin Bosse, Elisabeth Botelho-Nevers, Lila Bouadma, Olivier Bouchaud, Sabelline Bouchez, Dounia Bouhmani, Damien Bouhour, Kévin Bouiller, Laurence Bouillet, Camile Bouisse, Thipsavanh Bounphiengsy, Latsaniphone Bountthasavong, Anne-Sophie Boureau, John Bourke, Maude Bouscambert, Aurore Bousquet, Jason Bouziotis, Bianca Boxma, Marielle Boyer-Besseyre, Maria Boylan, Fernando Augusto Bozza, Axelle Braconnier, Cynthia Braga, Timo Brandenburger, Filipa Brás Monteiro, Luca Brazzi, Dorothy Breen, Patrick Breen, Kathy Brickell, Alex Browne, Shaunagh Browne, Nicolas Brozzi, Sonja Hjellegjerde Brunvoll, Marjolein Brusse-Keizer, Petra Bryda, Nina Buchtele, Polina Bugaeva, Marielle Buisson, Danilo Buonsenso, Erlina Burhan, Aidan Burrell, Ingrid G. Bustos, Denis Butnaru, André Cabie, Susana Cabral, Eder Caceres, Cyril Cadoz, Rui Caetano Garcês, Mia Callahan, Kate Calligy, Jose Andres Calvache, Caterina Caminiti, João Camões, Valentine Campana, Paul Campbell, Josie Campisi, Cecilia Canepa, Mireia Cantero, Janice Caoili, Pauline Caraux-Paz, Sheila Cárcel, Chiara Simona Cardellino, Filipa Cardoso, Filipe Cardoso, Nelson Cardoso, Sofia Cardoso, Simone Carelli, Francesca Carlacci, Nicolas Carlier, Thierry Carmoi, Gayle Carney, Inês Carqueja, Marie-Christine Carret, François Martin Carrier, Ida Carroll, Gail Carson, Leonor Carvalho, Maire-Laure Casanova, Mariana Cascão, Siobhan Casey, José Casimiro, Bailey Cassandra, Silvia Castañeda, Nidyanara Castanheira, Guylaine Castor-Alexandre, Ivo Castro, Ana Catarino, François-Xavier Catherine, Paolo Cattaneo, Roberta Cavalin, Giulio Giovanni Cavalli, Alexandros Cavayas, Adrian Ceccato, Shelby Cerkovnik, Minerva Cervantes-Gonzalez, Muge Cevik, Anissa Chair, Catherine Chakveatze, Bounthavy Chaleunphon, Adrienne Chan, Meera Chand, Christelle Chantalat Auger, Jean-Marc Chapplain, Charlotte Charpentier, Julie Chas, Allegra Chatterjee, Mobin Chaudry, Jonathan Samuel Chávez Iñiguez, Anjellica Chen, Yih-Sharng Chen, Léo Chenard, Matthew Pellan Cheng, Antoine Cheret, Alfredo Antonio Chetta, Thibault Chiarabini, Julian Chica, Suresh Kumar Chidambaram, Leong Chin Tho, Catherine Chirouze, Davide Chiumello, Hwa Jin Cho, Sung-Min Cho, Bernard Cholley, Danoy Chommanam, Marie-Charlotte Chopin, Ting Soo Chow, Yock Ping Chow, Nathaniel Christy, Hiu Jian Chua, Jonathan Chua, Jose Pedro Cidade, José Miguel Cisneros Herreros, Barbara Wanjiru Citarella, Anna Ciullo, Emma Clarke, Jennifer Clarke, Rolando Claure-Del Granado, Sara Clohisey, Perren J. Cobb, Cassidy Codan, Caitriona Cody, Alexandra Coelho, Megan Coles, Gwenhaël Colin, Michael Collins, Sebastiano Maria Colombo, Pamela Combs, Jennifer Connolly, Marie Connor, Anne Conrad, Sofía Contreras, Elaine Conway, Graham S. Cooke, Mary Copland, Hugues Cordel, Amanda Corley, Sabine Cornelis, Alexander Daniel Cornet, Arianne Joy Corpuz, Andrea Cortegiani, Grégory Corvaisier, Emma Costigan, Camille Couffignal, Sandrine Couffin-Cadiergues, Roxane Courtois, Stéphanie Cousse, Rachel Cregan, Charles Crepy D'Orleans, Cosimo Cristella, Sabine Croonen, Gloria Crowl, Jonathan Crump, Claudina Cruz, Juan Luis Cruz Bermúdez, Jaime Cruz Rojo, Marc Csete, Alberto Cucino, Ailbhe Cullen, Matthew Cummings, Ger Curley, Elodie Curlier, Colleen Curran, Paula Custodio, Ana da Silva Filipe, Charlene Da Silveira, Al-Awwab Dabaliz, Andrew Dagens, John Arne Dahl, Darren Dahly, Peter Daley, Heidi Dalton, Jo Dalton, Seamus Daly, Juliana Damas, Federico D'Amico, Nick Daneman, Corinne Daniel, Emmanuelle A Dankwa, Jorge Dantas, Frédérick D'Aragon, Mark de Boer, Menno de Jong, Gillian de Loughry, Diego de Mendoza, Etienne De Montmollin, Rafael Freitas de Oliveira França, Ana Isabel de Pinho Oliveira, Rosanna De Rosa, Cristina De Rose, Thushan de Silva, Peter de Vries, Jillian Deacon, David Dean, Alexa Debard, Bianca DeBenedictis, Marie-Pierre Debray, Nathalie DeCastro, William Dechert, Lauren Deconninck, Romain Decours, Eve Defous, Isabelle Delacroix, Eric Delaveuve, Karen Delavigne, Nathalie M. Delfos, Ionna Deligiannis, Andrea Dell'Amore, Christelle Delmas, Pierre Delobel, Corine Delsing, Elisa Demonchy, Emmanuelle Denis, Dominique Deplanque, Pieter Depuydt, Mehul Desai, Diane Descamps, Mathilde Desvallées, Santi Dewayanti, Pathik Dhanger, Alpha Diallo, Sylvain Diamantis, André Dias, Andrea Dias, Fernanda Dias Da Silva, Juan Jose Diaz, Priscila Diaz, Rodrigo Diaz, Kévin Didier, Jean-Luc Diehl, Wim Dieperink, Jérôme Dimet, Vincent Dinot, Fara Diop, Alphonsine Diouf, Yael Dishon, Félix Djossou, Annemarie B. Docherty, Helen Doherty, Arjen M Dondorp, Andy Dong, Christl A. Donnelly, Maria Donnelly, Chloe Donohue, Sean Donohue, Yoann Donohue, Peter Doran, Céline Dorival, Eric D'Ortenzio, Phouvieng Douangdala, James Joshua Douglas, Renee Douma, Nathalie Dournon, Triona Downer, Joanne Downey, Mark Downing, Tom Drake, Aoife Driscoll, Amiel A. Dror, Murray Dryden, Claudio Duarte Fonseca, Vincent Dubee, François Dubos, Audrey Dubot-Pérès, Alexandre Ducancelle, Toni Duculan, Susanne Dudman, Abhijit Duggal, Paul Dunand, Jake Dunning, Mathilde Duplaix, Emanuele Durante-Mangoni, Lucian Durham III, Bertrand Dussol, Juliette Duthoit, Xavier Duval, Anne Margarita Dyrhol-Riise, Sim Choon Ean, Marco Echeverria-Villalobos, Giorgio Economopoulos, Michael Edelstein, Siobhan Egan, Linn Margrete Eggesbø, Carla Eira, Mohammed El Sanharawi, Subbarao Elapavaluru, Brigitte Elharrar, Jacobien Ellerbroek, Merete Ellingjord-Dale, Philippine Eloy, Tarek Elshazly, Iqbal Elyazar, Isabelle Enderle, Tomoyuki Endo, Chan Chee Eng, Ilka Engelmann, Vincent Enouf, Olivier Epaulard, Martina Escher, Mariano Esperatti, Hélène Esperou, Catarina Espírito Santo, Marina Esposito-Farese, Lorinda Essuman, João Estevão, Manuel Etienne, Nadia Ettalhaoui, Anna Greti Everding, Mirjam Evers, Isabelle Fabre, Marc Fabre, Amna Faheem, Arabella Fahy, Cameron J. Fairfield, Zul Fakar, Komal Fareed, Pedro Faria, Ahmed Farooq, Hanan Fateena, Arie Zainul Fatoni, Karine Faure, Raphaël Favory, Mohamed Fayed, Niamh Feely, Laura Feeney, Jorge Fernandes, Marília Andreia Fernandes, Susana Fernandes, François-Xavier Ferrand, Eglantine Ferrand Devouge, Joana Ferrão, Carlo Ferrari, Mário Ferraz, Benigno Ferreira, Bernardo Ferreira, Isabel Ferreira, Sílvia Ferreira, Ricard Ferrer-Roca, Nicolas Ferriere, Céline Ficko, Claudia Figueiredo-Mello, William Finlayson, Juan Fiorda, Thomas Flament, Clara Flateau, Tom Fletcher, Aline-Marie Florence, Letizia Lucia Florio, Brigid Flynn, Deirdre Flynn, Federica Fogliazza, Claire Foley, Jean Foley, Victor Fomin, Tatiana Fonseca, Patricia Fontela, Karen Forrest, Simon Forsyth, Denise Foster, Giuseppe Foti, Erwan Fourn, Robert A. Fowler, Marianne Fraher, Diego Franch-Llasat, Christophe Fraser, John F Fraser, Marcela Vieira Freire, Ana Freitas Ribeiro, Craig French, Caren Friedrich, Ricardo Fritz, Stéphanie Fry, Nora Fuentes, Masahiro Fukuda, Argin G, Valérie Gaborieau, Rostane Gaci, Massimo Gagliardi, Jean-Charles Gagnard, Nathalie Gagné, Amandine Gagneux-Brunon, Sérgio Gaião, Linda Gail Skeie, Phil Gallagher, Elena Gallego Curto, Carrol Gamble, Yasmin Gani, Arthur Garan, Rebekha Garcia, Noelia García Barrio, Julia Garcia-Diaz, Esteban Garcia-Gallo, Navya Garimella, Federica Garofalo, Denis Garot, Valérie Garrait, Basanta Gauli, Nathalie Gault, Aisling Gavin, Anatoliy Gavrylov, Alexandre Gaymard, Johannes Gebauer, Eva Geraud, Louis Gerbaud Morlaes, Nuno Germano, praveen kumar ghisulal, Jade Ghosn, Marco Giani, Carlo Giaquinto, Jess Gibson, Tristan Gigante, Morgane Gilg, Elaine Gilroy, Guillermo Giordano, Michelle Girvan, Valérie Gissot, Jesse Gitaka, Gezy Giwangkancana, Daniel Glikman, Petr Glybochko, Eric Gnall, Geraldine Goco, François Goehringer, Siri Goepel, Jean-Christophe Goffard, Jin Yi Goh, Jonathan Golob, Rui Gomes, Kyle Gomez, Joan Gómez-Junyent, Marie Gominet, Bronner P. Gonçalves, Alicia Gonzalez, Patricia Gordon, Yanay Gorelik, Isabelle Gorenne, Conor Gormley, Laure Goubert, Cécile Goujard, Tiphaine Goulenok, Margarite Grable, Jeronimo Graf, Edward Wilson Grandin, Pascal Granier, Giacomo Grasselli, Lorenzo Grazioli, Christopher A. Green, Courtney Greene, William Greenhalf, Segolène Greffe, Domenico Luca Grieco, Matthew Griffee, Fiona Griffiths, Ioana Grigoras, Albert Groenendijk, Anja Grosse Lordemann, Heidi Gruner, Yusing Gu, Fabio Guarracino, Jérémie Guedj, Martin Guego, Dewi Guellec, Anne-Marie Guerguerian, Daniela Guerreiro, Romain Guery, Anne Guillaumot, Laurent Guilleminault, Maisa Guimarães de Castro, Thomas Guimard, Marieke Haalboom, Daniel Haber, Hannah Habraken, Ali Hachemi, Amy Hackmann, Nadir Hadri, Fakhir Haidri, Sheeba Hakak, Adam Hall, Matthew Hall, Sophie Halpin, Jawad Hameed, Ansley Hamer, Rebecca Hamidfar, Bato Hammarström, Terese Hammond, Lim Yuen Han, Rashan Haniffa, Kok Wei Hao, Hayley Hardwick, Ewen M. Harrison, Janet Harrison, Samuel Bernard Ekow Harrison, Alan Hartman, Mohd Shahnaz Hasan, Junaid Hashmi, Muhammad Hayat, Ailbhe Hayes, Leanne Hays, Jan Heerman, Lars Heggelund, Ross Hendry, Martina Hennessy, Aquiles Henriquez-Trujillo, Maxime Hentzien, Diana  Hernandez, Jaime Hernandez-Montfort, Daniel Herr, Andrew Hershey, Liv Hesstvedt, Astarini Hidayah, Dawn Higgins, Eibhlin Higgins, Rupert Higgins, Rita Hinchion, Samuel Hinton, Hiroaki Hiraiwa, Hikombo Hitoto, Antonia Ho, Yi Bin Ho, Alexandre Hoctin, Isabelle Hoffmann, Wei Han Hoh, Oscar Hoiting, Rebecca Holt, Jan Cato Holter, Peter Horby, Juan Pablo Horcajada, Koji Hoshino, Kota Hoshino, Ikram Houas, Catherine L. Hough, Stuart Houltham, Jimmy Ming-Yang Hsu, Jean-Sébastien Hulot, Stella Huo, Abby Hurd, Iqbal Hussain, Samreen Ijaz, Arfan Ikram, Hajnal-Gabriela Illes, Patrick Imbert, Mohammad Imran, Rana Imran Sikander, Aftab Imtiaz, Hugo Inácio, Carmen Infante Dominguez, Yun Sii Ing, Elias Iosifidis, Mariachiara Ippolito, Vera Irawany, Sarah Isgett, Tiago Isidoro, Nadiah Ismail, Margaux Isnard, Mette Stausland Istre, Junji Itai, Asami Ito, Daniel Ivulich, Danielle Jaafar, Salma Jaafoura, Julien Jabot, Clare Jackson, Nina Jamieson, Victoria Janes, Pierre Jaquet, Waasila Jassat, Coline Jaud-Fischer, Stéphane Jaureguiberry, Jeffrey Javidfar, Denise Jaworsky, Florence Jego, Anilawati Mat Jelani, Synne Jenum, Ruth Jimbo-Sotomayor, Ong Yiaw Joe, Ruth N. Jorge García, Silje Bakken Jørgensen, Cédric Joseph, Mark Joseph, Swosti Joshi, Mercé Jourdain, Philippe Jouvet, Jennifer June, Anna Jung, Hanna Jung, Dafsah Juzar, Ouifiya Kafif, Florentia Kaguelidou, Neerusha Kaisbain, Thavamany Kaleesvran, Sabina Kali, Alina Kalicinska, Karl Trygve Kalleberg, Smaragdi Kalomoiri, Muhammad Aisar Ayadi Kamaluddin, Zul Amali Che Kamaruddin, Nadiah Kamarudin, Kavita Kamineni, Darshana Hewa Kandamby, Chris Kandel, Kong Yeow Kang, Darakhshan Kanwal, Dyah Kanyawati, Pratap Karpayah, Todd Karsies, Christiana Kartsonaki, Daisuke Kasugai, Anant Kataria, Kevin Katz, Aasmine Kaur, Tatsuya Kawasaki, Christy Kay, Lamees Kayyali, Hannah Keane, Seán Keating, Andrea Kelly, Aoife Kelly, Claire Kelly, Niamh Kelly, Sadie Kelly, Yvelynne Kelly, Maeve Kelsey, Ryan Kennedy, Kalynn Kennon, Sommay Keomany, Maeve Kernan, Younes Kerroumi, Sharma Keshav, Evelyne Kestelyn, Imrana Khalid, Osama Khalid, Antoine Khalil, Coralie Khan, Irfan Khan, Quratul Ain Khan, Sushil Khanal, Abid Khatak, Amin Khawaja, Michelle E Kho, Denisa Khoo, Ryan Khoo, Saye Khoo, Nasir Khoso, Khor How Kiat, Yuri Kida, Harrison Kihuga, Peter Kiiza, Beathe Kiland Granerud, Anders Benjamin Kildal, Jae Burm Kim, Antoine Kimmoun, Detlef Kindgen-Milles, Alexander King, Nobuya Kitamura, Eyrun Floerecke Kjetland Kjetland, Paul Klenerman, Rob Klont, Gry Kloumann Bekken, Stephen R Knight, Robin Kobbe, Paa Kobina Forson, Chamira Kodippily, Malte Kohns Vasconcelos, Sabin Koirala, Mamoru Komatsu, Franklina Korkor Abebrese, Volkan Korten, Caroline Kosgei, Arsène Kpangon, Karolina Krawczyk, Sudhir Krishnan, Vinothini Krishnan, Oksana Kruglova, Deepali Kumar, Ganesh Kumar, Mukesh Kumar, Bharath Kumar Tirupakuzhi Vijayaraghavan, Pavan Kumar Vecham, Dinesh Kuriakose, Ethan Kurtzman, Neurinda Permata Kusumastuti, Demetrios Kutsogiannis, Galyna Kutsyna, Ama Kwakyewaa Bedu-Addo, Konstantinos Kyriakoulis, Raph L. Hamers, Marie Lachatre, Marie Lacoste, John G. Laffey, Nadhem Lafhej, Marie Lagrange, Fabrice Laine, Olivier Lairez, Sanjay Lakhey, Antonio Lalueza, Marc Lambert, François Lamontagne, Marie Langelot-Richard, Vincent Langlois, Eka Yudha Lantang, Marina Lanza, Cédric Laouénan, Samira Laribi, Delphine Lariviere, Stéphane Lasry, Naveed Latif, Odile Launay, Didier Laureillard, Yoan Lavie-Badie, Andrew Law, Cassie Lawrence, Teresa Lawrence, Minh Le, Clément Le Bihan, Cyril Le Bris, Georges Le Falher, Lucie Le Fevre, Quentin Le Hingrat, Marion Le Maréchal, Soizic Le Mestre, Gwenaël Le Moal, Vincent Le Moing, Hervé Le Nagard, Paul Le Turnier, Ema Leal, Marta Leal Santos, Biing Horng Lee, Heng Gee Lee, James Lee, Jennifer Lee, Su Hwan Lee, Todd C. Lee, Yi Lin Lee, Gary Leeming, Bénédicte Lefebvre, Laurent Lefebvre, Benjamin Lefèvre, Sylvie LeGac, Jean-Daniel Lelievre, François Lellouche, Adrien Lemaignen, Véronique Lemee, Anthony Lemeur, Gretchen Lemmink, Ha Sha Lene, Jenny Lennon, Rafael León, Marc Leone, Michela Leone, François-Xavier Lescure, Olivier Lesens, Mathieu Lesouhaitier, Amy Lester-Grant, Andrew Letizia, Sophie Letrou, Bruno Levy, Yves Levy, Claire Levy-Marchal, Katarzyna Lewandowska, Erwan L'Her, Gianluigi Li Bassi, Janet Liang, Ali Liaquat, Geoffrey Liegeon, Kah Chuan Lim, Wei Shen Lim, Chantre Lima, Bruno Lina, Lim Lina, Andreas Lind, Maja Katherine Lingad, Guillaume Lingas, Sylvie Lion-Daolio, Samantha Lissauer, Keibun Liu, Marine Livrozet, Patricia Lizotte, Antonio Loforte, Navy Lolong, Leong Chee Loon, Diogo Lopes, Dalia Lopez-Colon, Jose W. Lopez-Revilla, Anthony L. Loschner, Paul Loubet, Bouchra Loufti, Guillame Louis, Silvia Lourenco, Lara Lovelace-Macon, Lee Lee Low, Marije Lowik, Jia Shyi Loy, Jean Christophe Lucet, Carlos Lumbreras Bermejo, Carlos M. Luna, Olguta Lungu, Liem Luong, Nestor Luque, Dominique Luton, Nilar Lwin, Ruth Lyons, Olavi Maasikas, Oryane Mabiala, Sarah MacDonald, Moïse Machado, Sara Machado, Gabriel Macheda, Juan Macias Sanchez, Jai Madhok, Hashmi Madiha, Guillermo Maestro de la Calle, Jacob Magara, Giuseppe Maglietta, Rafael Mahieu, Sophie Mahy, Ana Raquel Maia, Lars S. Maier, Mylène Maillet, Thomas Maitre, Maria Majori, Maximilian Malfertheiner, Nadia Malik, Paddy Mallon, Fernando Maltez, Denis Malvy, Patrizia Mammi, Victoria Manda, Jose M. Mandei, Laurent Mandelbrot, Frank Manetta, Julie Mankikian, Edmund Manning, Aldric Manuel, Ceila Maria Sant`Ana Malaque, Daniel Marino, Flávio Marino, Samuel Markowicz, Charbel Maroun Eid, Ana Marques, Catherine Marquis, Brian Marsh, Laura Marsh, Megan Marshal, John Marshall, Celina Turchi Martelli, Dori-Ann Martin, Emily Martin, Guillaume Martin-Blondel, Alessandra Martinelli, Ignacio Martin-Loeches, Martin Martinot, Alejandro Martín-Quiros, Ana Martins, João Martins, Nuno Martins, Caroline Martins Rego, Gennaro Martucci, Olga Martynenko, Eva Miranda Marwali, Marsilla Marzukie, Juan Fernado Masa Jimenez, David Maslove, Phillip Mason, Sabina Mason, Sobia Masood, Basri Mat Nor, Moshe Matan, Henrique Mateus Fernandes, Meghena Mathew, Daniel Mathieu, Mathieu Mattei, Romans Matulevics, Laurence Maulin, Michael Maxwell, Javier Maynar, Mayfong Mayxay, Thierry Mazzoni, Lisa Mc Sweeney, Colin McArthur, Aine McCarthy, Anne McCarthy, Colin McCloskey, Rachael McConnochie, Sherry McDermott, Sarah E. McDonald, Aine McElroy, Samuel McElwee, Victoria McEneany, Natalie McEvoy, Allison McGeer, Chris McKay, Johnny McKeown, Kenneth A. McLean, Paul McNally, Bairbre McNicholas, Elaine McPartlan, Edel Meaney, Cécile Mear-Passard, Maggie Mechlin, Maqsood Meher, Omar Mehkri, Ferruccio Mele, Luis Melo, Kashif Memon, Joao Joao Mendes, Ogechukwu Menkiti, Kusum Menon, France Mentré, Alexander J. Mentzer, Emmanuelle Mercier, Noémie Mercier, Antoine Merckx, Mayka Mergeay-Fabre, Blake Mergler, Laura Merson, Tiziana Meschi, António Mesquita, Roberta Meta, Osama Metwally, Agnès Meybeck, Dan Meyer, Alison M. Meynert, Vanina Meysonnier, Amina Meziane, Mehdi Mezidi, Giuliano Michelagnoli, Céline Michelanglei, Isabelle Michelet, Efstathia Mihelis, Vladislav Mihnovit, Hugo Miranda-Maldonado, Nor Arisah Misnan, Nik Nur Eliza Mohamed, Tahira Jamal Mohamed, Asma Moin, Elena Molinos, Brenda Molloy, Sinead Monahan, Mary Mone, Agostinho Monteiro, Claudia Montes, Giorgia Montrucchio, Sarah Moore, Shona C. Moore, Lina Morales Cely, Lucia Moro, Diego Rolando Morocho Tutillo, Ben Morton, Catherine Motherway, Ana Motos, Hugo Mouquet, Clara Mouton Perrot, Julien Moyet, Caroline Mudara, Aisha Kalsoom Mufti, Ng Yong Muh, Dzawani Muhamad, Jimmy Mullaert, Fredrik Müller, Karl Erik Müller, Daniel Munblit, Syed Muneeb, Nadeem Munir, Laveena Munshi, Aisling Murphy, Lorna Murphy, Patrick Murray, Marlène Murris, Srinivas Murthy, Himed Musaab, Alamin Mustafa, Carlotta Mutti, Himasha Muvindi, Gugapriyaa Muyandy, Dimitra Melia Myrodia, Farah Nadia Mohd-Hanafiah, Dave Nagpal, Alex Nagrebetsky, Mangala Narasimhan, Nageswaran Narayanan, Rashid Nasim Khan, Alasdair Nazerali-Maitland, Nadège Neant, Holger Neb, Coca Necsoi, Nikita Nekliudov, Matthew Nelder, Erni Nelwan, Raul Neto, Emily Neumann, Bernardo Neves, Pauline Yeung Ng, Anthony Nghi, Jane Ngure, Duc Nguyen, Orna Ni Choileain, Niamh Ni Leathlobhair, Alistair Nichol, Prompak Nitayavardhana, Stephanie Nonas, Nurul Amani Mohd Noordin, Marion Noret, Nurul Faten Izzati Norharizam, Lisa Norman, Anita North, Alessandra Notari, Mahdad Noursadeghi, Karolina Nowicka, Adam Nowinski, Saad Nseir, Jose I Nunez, Nurnaningsih Nurnaningsih, Dwi Utomo Nusantara, Elsa Nyamankolly, Anders Benteson Nygaard, Fionnuala O Brien, Annmarie O Callaghan, Annmarie O'Callaghan, Giovanna Occhipinti, Derbrenn OConnor, Max O'Donnell, Tawnya Ogston, Takayuki Ogura, Tak-Hyuk Oh, Sophie O'Halloran, Katie O'Hearn, Shinichiro Ohshimo, Agnieszka Oldakowska, João Oliveira, Larissa Oliveira, Piero L. Olliaro, Conar O'Neil, David S.Y. Ong, Jee Yan Ong, Wilna Oosthuyzen, Anne Opavsky, Peter Openshaw, Saijad Orakzai, Claudia Milena Orozco-Chamorro, Andrés Orquera, Jamel Ortoleva, Javier Osatnik, Linda O'Shea, Miriam O'Sullivan, Siti Zubaidah Othman, Paul Otiku, Nadia Ouamara, Rachida Ouissa, Clark Owyang, Eric Oziol, Maïder Pagadoy, Justine Pages, Amanda Palacios, Massimo Palmarini, Giovanna Panarello, Prasan Kumar Panda, Hem Paneru, Lai Hui Pang, Mauro Panigada, Nathalie Pansu, Aurélie Papadopoulos, Paolo Parducci, Edwin Fernando Paredes Oña, Rachael Parke, Melissa Parker, Vieri Parrini, Taha Pasha, Jérémie Pasquier, Bruno Pastene, Fabian Patauner, Mohan Dass Pathmanathan, Luís Patrão, Patricia Patricio, Juliette Patrier, Laura Patrizi, Lisa Patterson, Rajyabardhan Pattnaik, Christelle Paul, Mical Paul, Jorge Paulos, William A. Paxton, Jean-François Payen, Kalaiarasu Peariasamy, Miguel Pedrera Jiménez, Giles J. Peek, Florent Peelman, Nathan Peiffer-Smadja, Vincent Peigne, Mare Pejkovska, Paolo Pelosi, Ithan D. Peltan, Rui Pereira, Daniel Perez, Luis Periel, Thomas Perpoint, Antonio Pesenti, Vincent Pestre, Lenka Petrou, Michele Petrovic, Ventzislava Petrov-Sanchez, Frank Olav Pettersen, Gilles Peytavin, Scott Pharand, Ooyanong Phonemixay, Soulichanya Phoutthavong, Michael Piagnerelli, Walter Picard, Olivier Picone, Maria de Piero, Carola Pierobon, Djura Piersma, Carlos Pimentel, Raquel Pinto, Valentine Piquard, Catarina Pires, Isabelle Pironneau, Lionel Piroth, Roberta Pisi, Ayodhia Pitaloka, Riinu Pius, Simone Piva, Laurent Plantier, Hon Shen Png, Julien Poissy, Ryadh Pokeerbux, Maria Pokorska-Spiewak, Sergio Poli, Georgios Pollakis, Diane Ponscarme, Jolanta Popielska, Diego Bastos Porto, Andra-Maris Post, Douwe F. Postma, Pedro Povoa, Diana Póvoas, Jeff Powis, Sofia Prapa, Viladeth Praphasiri, Sébastien Preau, Christian Prebensen, Jean-Charles Preiser, Anton Prinssen, Mark G. Pritchard, Gamage Dona Dilanthi Priyadarshani, Lucia Proença, Sravya Pudota, Oriane Puéchal, Bambang Pujo Semedi, Mathew Pulicken, Matteo Puntoni, Gregory Purcell, Luisa Quesada, Vilmaris Quinones-Cardona, Víctor Quirós González, Else Quist-Paulsen, Mohammed Quraishi, Fadi-Fadi Qutishat, Maia Rabaa, Christian Rabaud, Ebenezer Rabindrarajan, Aldo Rafael, Marie Rafiq, Gabrielle Ragazzo, Mutia Rahardjani, Ahmad Kashfi Haji Ab Rahman, Rozanah Abd Rahman, Arsalan Rahutullah, Fernando Rainieri, Giri Shan Rajahram, Pratheema Ramachandran, Nagarajan Ramakrishnan, José Ramalho, Kollengode Ramanathan, Ahmad Afiq Ramli, Blandine Rammaert, Grazielle Viana Ramos, Anais Rampello, Asim Rana, Rajavardhan Rangappa, Ritika Ranjan, Elena Ranza, Christophe Rapp, Aasiyah Rashan, Thalha Rashan, Ghulam Rasheed, Menaldi Rasmin, Indrek Rätsep, Cornelius Rau, Francesco Rausa, Tharmini Ravi, Ali Raza, Andre Real, Stanislas Rebaudet, Sarah Redl, Brenda Reeve, Attaur Rehman, Liadain Reid, Liadain Reid, Dag Henrik Reikvam, Renato Reis, Jordi Rello, Jonathan Remppis, Martine Remy, Hongru Ren, Hanna Renk, Anne-Sophie Resseguier, Matthieu Revest, Oleksa Rewa, Luis Felipe Reyes, Tiago Reyes, Maria Ines Ribeiro, Antonia Ricchiuto, David Richardson, Denise Richardson, Laurent Richier, Siti Nurul Atikah Ahmad Ridzuan, Jordi Riera, Ana L Rios, Asgar Rishu, Patrick Rispal, Karine Risso, Maria Angelica Rivera Nuñez, Nicholas Rizer, Doug Robb, Chiara Robba, André Roberto, Stephanie Roberts, David L. Robertson, Olivier Robineau, Ferran Roche-Campo, Paola Rodari, Simão Rodeia, Julia Rodriguez Abreu, Bernhard Roessler, Claire Roger, Pierre-Marie Roger, Emmanuel Roilides, Amanda Rojek, Juliette Romaru, Roberto Roncon-Albuquerque Jr, Mélanie Roriz, Manuel Rosa-Calatrava, Michael Rose, Dorothea Rosenberger, Andrea Rossanese, Matteo Rossetti, Sandra Rossi, Bénédicte Rossignol, Patrick Rossignol, Stella Rousset, Carine Roy, Benoît Roze, Desy Rusmawatiningtyas, Clark D. Russell, Maeve Ryan, Maria Ryan, Steffi Ryckaert, Aleksander Rygh Holten, Isabela Saba, Luca Sacchelli, Sairah Sadaf, Musharaf Sadat, Valla Sahraei, Nadia Saidani, Maximilien Saint-Gilles, Pranya Sakiyalak, Nawal Salahuddin, Leonardo Salazar, Jodat Saleem, Nazal Saleh, Gabriele Sales, Stéphane Sallaberry, Charlotte Salmon Gandonniere, Hélène Salvator, Olivier Sanchez, Xavier Sánchez Choez, Kizy Sanchez de Oliveira, Angel Sanchez-Miralles, Vanessa Sancho-Shimizu, Gyan Sandhu, Zulfiqar Sandhu, Pierre-François Sandrine, Oana Sandulescu, Marlene Santos, Shirley Sarfo-Mensah, Bruno Sarmento Banheiro, Iam Claire E. Sarmiento, Benjamine Sarton, Sree Satyapriya, Rumaisah Satyawati, Egle Saviciute, Parthena Savvidou, Yen Tsen Saw, Justin Schaffer, Tjard Schermer, Arnaud Scherpereel, Marion Schneider, Stephan Schroll, Michael Schwameis, Gary Schwartz, Brendan Scicluna, Janet T. Scott, James Scott-Brown, Nicholas Sedillot, Tamara Seitz, Jaganathan Selvanayagam, Mageswari Selvarajoo, Caroline Semaille, Malcolm G. Semple, Rasidah Bt Senian, Eric Senneville, Claudia Sepulveda, Filipa Sequeira, Tânia Sequeira, Ary Serpa Neto, Pablo Serrano Balazote, Ellen Shadowitz, Syamin Asyraf Shahidan, Mohammad Shamsah, Anuraj Shankar, Shaikh Sharjeel, Pratima Sharma, Catherine A. Shaw, Victoria Shaw, John Robert Sheenan, Ashraf Sheharyar, Dr. Rajesh Mohan Shetty, Haixia Shi, Nisreen Shiban, Mohiuddin Shiekh, Takuya Shiga, Nobuaki Shime, Hiroaki Shimizu, Keiki Shimizu, Naoki Shimizu, Sally Shrapnel, Pramesh Sundar Shrestha, Shubha Kalyan Shrestha, Hoi Ping Shum, Nassima Si Mohammed, Ng Yong Siang, Moses Siaw-Frimpong, Jeanne Sibiude, Bountoy Sibounheuang, Atif Siddiqui, Louise Sigfrid, Piret Sillaots, Catarina Silva, Maria Joao Silva, Rogério Silva, Benedict Sim Lim Heng, Wai Ching Sin, Dario Sinatti, Budha Charan Singh, Punam Singh, Pompini Agustina Sitompul, Karisha Sivam, Vegard Skogen, Sue Smith, Benjamin Smood, Coilin Smyth, Michelle Smyth, Morgane Snacken, Dominic So, Tze Vee Soh, Lene Bergendal Solberg, Joshua Solomon, Tom Solomon, Emily Somers, Agnès Sommet, Myung Jin Song, Rima Song, Tae Song, Jack Song Chia, Michael Sonntagbauer, Azlan Mat Soom, Arne Søraas, Camilla Lund Søraas, Albert Sotto, Edouard Soum, Ana Chora Sousa, Marta Sousa, Maria Sousa Uva, Vicente Souza-Dantas, Alexandra Sperry, Elisabetta Spinuzza, B. P. Sanka Ruwan Sri Darshana, Shiranee Sriskandan, Sarah Stabler, Thomas Staudinger, Stephanie-Susanne Stecher, Trude Steinsvik, Ymkje Stienstra, Birgitte Stiksrud, Eva Stolz, Amy Stone, Adrian Streinu-Cercel, Anca Streinu-Cercel, Ami Stuart, David Stuart, Richa Su, Decy Subekti, Gabriel Suen, Jacky Y. Suen, Prasanth Sukumar, Asfia Sultana, Charlotte Summers, Dubravka Supic, Deepashankari Suppiah, Magdalena Surovcová, Atie Suwarti, Andrey Svistunov, Sarah Syahrin, Konstantinos Syrigos, Jaques Sztajnbok, Konstanty Szuldrzynski, Shirin Tabrizi, Fabio S. Taccone, Lysa Tagherset, Shahdattul Mawarni Taib, Ewa Talarek, Sara Taleb, Jelmer Talsma, Renaud Tamisier, Maria Lawrensia Tampubolon, Kim Keat Tan, Le Van Tan, Yan Chyi Tan, Clarice Tanaka, Hiroyuki Tanaka, Taku Tanaka, Hayato Taniguchi, Huda Taqdees, Arshad Taqi, Coralie Tardivon, Pierre Tattevin, M Azhari Taufik, Hassan Tawfik, Richard S. Tedder, Tze Yuan Tee, João Teixeira, Sofia Tejada, Marie-Capucine Tellier, Sze Kye Teoh, Vanessa Teotonio, François Téoulé, Pleun Terpstra, Olivier Terrier, Nicolas Terzi, Hubert Tessier-Grenier, Adrian Tey, Alif Adlan Mohd Thabit, Anand Thakur, Zhang Duan Tham, Suvintheran Thangavelu, Elmi Theron, Vincent Thibault, Simon-Djamel Thiberville, Benoît Thill, Jananee Thirumanickam, Shaun Thompson, David Thomson, Emma C. Thomson, Surain Raaj Thanga Thurai, Duong Bich Thuy, Ryan S. Thwaites, Andrea Ticinesi, Paul Tierney, Vadim Tieroshyn, Peter S Timashev, Jean-François Timsit, Noémie Tissot, Fiona Toal, Jordan Zhien Yang Toh, Maria Toki, Kristian Tonby, Sia Loong Tonnii, Marta Torre, Antoni Torres, Margarida Torres, Rosario Maria Torres Santos-Olmo, Hernando Torres-Zevallos, Michael Towers, Tony Trapani, Huynh Trung Trieu, Théo Trioux, Cécile Tromeur, Ioannis Trontzas, Tiffany Trouillon, Jeanne Truong, Christelle Tual, Sarah Tubiana, Helen Tuite, Jean-Marie Turmel, Lance C.W. Turtle, Anders Tveita, Pawel Twardowski, Makoto Uchiyama, PG Ishara Udayanga, Andrew Udy, Roman Ullrich, Alberto Uribe, Asad Usman, Timothy M. Uyeki, Cristinava Vajdovics, Piero Valentini, Luís Val-Flores, Ana Luiza Valle, Amélie Valran, Ilaria Valzano, Stijn Van de Velde, Marcel van den Berge, Machteld Van der Feltz, Job van der Palen, Paul van der Valk, Nicky Van Der Vekens, Peter Van der Voort, Sylvie Van Der Werf, Marlice van Dyk, Laura van Gulik, Jarne Van Hattem, Carolien van Netten, Frank van Someren Greve, Gitte Van Twillert, Ilonka van Veen, Hugo Van Willigen, Noémie Vanel, Henk Vanoverschelde, Pooja Varghese, Michael Varrone, Shoban Raj Vasudayan, Charline Vauchy, Shaminee Veeran, Aurélie Veislinger, Sebastian Vencken, Sara Ventura, Annelies Verbon, James Vickers, José Ernesto Vidal, César Vieira, Deepak Vijayan, Joy Ann Villanueva, Judit Villar, Pierre-Marc Villeneuve, Andrea Villoldo, Nguyen Van Vinh Chau, Benoit Visseaux, Hannah Visser, Chiara Vitiello, Manivanh Vongsouvath, Harald Vonkeman, Fanny Vuotto, Noor Hidayu Wahab, Suhaila Abdul Wahab, Nadirah Abdul Wahid, Marina Wainstein, Laura Walsh, Wan Fadzlina Wan Muhd Shukeri, Chih-Hsien Wang, Steve Webb, Jia Wei, Katharina Weil, Tan Pei Wen, Sanne Wesselius, T. Eoin West, Murray Wham, Bryan Whelan, Nicole White, Paul Henri Wicky, Aurélie Wiedemann, Surya Otto Wijaya, Keith Wille, Suzette Willems, Virginie Williams, Evert-Jan Wils, Ng Wing Yiu, Calvin Wong, Teck Fung Wong, Xin Ci Wong, Yew Sing Wong, Natalie Wright, Gan Ee Xian, Lim Saio Xian, Kuan Pei Xuan, Ioannis Xynogalas, Sophie Yacoub, Siti Rohani Binti Mohd Yakop, Masaki Yamazaki, Yazdan Yazdanpanah, Nicholas Yee Liang Hing, Cécile Yelnik, Chian Hui Yeoh, Stephanie Yerkovich, Touxiong Yiaye, Toshiki Yokoyama, Hodane Yonis, Obada Yousif, Saptadi Yuliarto, Akram Zaaqoq, Marion Zabbe, Gustavo E Zabert, Kai Zacharowski, Masliza Zahid, Maram Zahran, Nor Zaila Binti Zaidan, Maria Zambon, Miguel Zambrano, Alberto Zanella, Konrad Zawadka, Nurul Zaynah, Hiba Zayyad, Alexander Zoufaly, David Zucman, Mazankowski Heart Institute.

\end{document}